%% file: NeuralVoxelRenderer.tex
\documentclass[10pt,twocolumn,letterpaper]{article}

\usepackage{cvpr}
\usepackage{times}
\usepackage{epsfig}
\usepackage{graphicx}
\usepackage{amsmath}
\usepackage{amssymb}
\usepackage{capt-of,etoolbox}

\input{our-commands}

\usepackage[pagebackref=true,breaklinks=true,letterpaper=true,colorlinks,bookmarks=false]{hyperref}
\usepackage{wrapfig}

\cvprfinalcopy % *** Uncomment this line for the final submission

\def\cvprPaperID{1541} % *** Enter the CVPR Paper ID here

\ifcvprfinal\fi
\begin{document}

%%%%%%%%% TITLE
\title{Neural Voxel Renderer: Learning an Accurate and Controllable Rendering Tool}

\author{Konstantinos Rematas\qquad Vittorio Ferrari\\Google Research}

% \maketitle
\thispagestyle{empty}
\twocolumn[{%
\renewcommand\twocolumn[1][]{#1}%
\maketitle
\vspace{-2mm}
\begin{center}
    \centering
    \includegraphics[width=.95\textwidth]{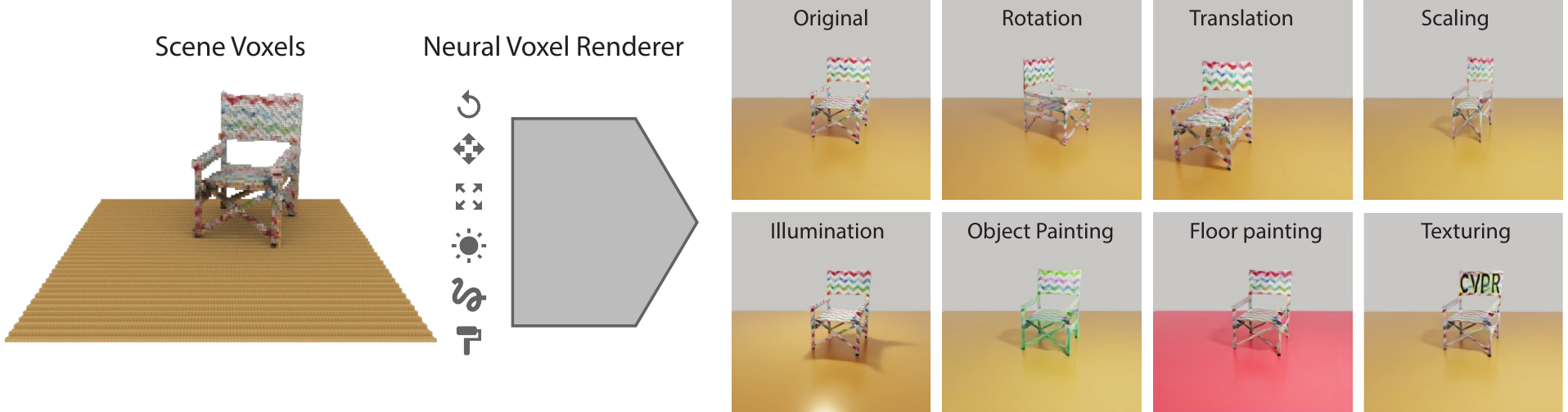}
    \captionof{figure}{Neural voxel renderer converts a set of colored voxels into a realistic and detailed image. It also allows elaborate modifications in the geometry or the appearance of the input that are faithfully represented in the synthesized image.}
    \label{fig:teaser}
\end{center}%
}]

\newcommand{\icol}[1]{% inline column vector
  \left(\begin{smallmatrix}#1\end{smallmatrix}\right)%
}

%%%%%%%%% ABSTRACT
\begin{abstract}
\vspace{-2mm}
We present a neural rendering framework that maps a voxelized scene into a high quality image. Highly-textured objects and scene element interactions are realistically rendered by our method, despite having a rough representation as an input.  Moreover, our approach allows controllable rendering: geometric and appearance modifications in the input are accurately propagated to the output. The user can move, rotate and scale an object, change its appearance and texture or modify the position of the light and all these edits are represented in the final rendering. 
We demonstrate the effectiveness of our approach by rendering scenes with varying appearance, from single color per object to complex, high-frequency textures. We show that our rerendering network can generate very detailed images that represent precisely the appearance of the input scene. 
Our experiments illustrate that our approach achieves more accurate image synthesis results compared to alternatives and can also handle low voxel grid resolutions. Finally, we show how our neural rendering framework can capture and faithfully render objects from real images and from a diverse set of classes.

\end{abstract}
%%%%%%%%% BODY TEXT

\input{1b_Intro}

\input{2_Related}

\input{3b_Method}
\input{4b_Experiments}

\input{5_Conclusion}

{\small
\bibliographystyle{ieee_fullname}
\bibliography{kostas}
}

\end{document}

%% file: our-commands.tex
\usepackage{soul}
\usepackage{xcolor}

\newcommand{\mypar}[1]{\vspace{1mm}\noindent{\bf #1}}

\def\figurePath{figures/}
\def\myfigure#1#2{\begin{figure}[t]\centering\includegraphics*[width = \linewidth]{\figurePath#1}\caption{#2}\label{fig:#1}\vspace{-5pt}\end{figure}}
\def\mycfigure#1#2{\begin{figure*}[ht]\centering\includegraphics*[clip, width = \linewidth]{\figurePath#1}\caption{#2}\label{fig:#1}\vspace{-6pt}\end{figure*}}

\def\myfigurehere#1#2{\begin{figure}[h]\centering\includegraphics*[width = \linewidth]{\figurePath#1}\caption{#2}\label{fig:#1}\vspace{-8pt}\end{figure}}

\newcommand{\refSec}[1]{Sec.~\ref{sec:#1}}
\newcommand{\refFig}[1]{Fig.~\ref{fig:#1}}
\newcommand{\refEq}[1]{Eq.~\eqref{eq:#1}}
\newcommand{\refTbl}[1]{Table~\ref{tbl:#1}}

\newcommand{\R}{\mathbb{R}}

\definecolor{unsurecolor}{rgb}{1,.85,.7}
\definecolor{changedcolor}{rgb}{.85,1,.7}

\soulregister\ref7
\soulregister\cite7
\soulregister\citeetal7
\soulregister\etal0
\soulregister\eg0
\soulregister\scene1
\soulregister\refTbl1

\DeclareGraphicsExtensions{.pdf,.ai,.psd,.jpg}
\newif\ifdraft
\drafttrue
\ifdraft
   \newcommand{\kostas}[1]{{\color{blue} {[\bf ko: #1]}}}
   \newcommand{\vitto}[1]{{\color{red} {[vt: #1]}}}
\else
  \newcommand{\kostas}[1]{}
   \newcommand{\vitto}[1]{}
\fi

%% file: 1b_Intro.tex
\vspace{-5mm}
\section{Introduction}
\vspace{-2mm}
What is the typical process for rendering a synthetic scene? In a 3D graphics software, like Blender~\cite{blender} and 3D Studio Max~\cite{3dsmax}, the user creates a set of geometric objects in a virtual 3D world, edits their material properties and adds the light sources. Once the desired configuration is achieved, the program renders the 3D scene into an image using a rendering algorithm such as Path Tracing~\cite{Kajiya}. While this setup unfolds the creativity of the user, it increases the learning complexity of the system, requires a lot of manual input and it is not differentiable.

The emergence of deep generative models introduced a new image synthesis medium. Generative adversarial networks~\cite{NIPS2014_5423} are able to produce highly realistic images of faces~\cite{Karras2019cvpr} or Imagenet~\cite{ILSVRC15} categories~\cite{brock2018large} using only class labels as input. Moreover, image-to-image translation~\cite{pix2pix2017} presented a framework where an input image featuring only partial information (\eg only edges), can be transformed into a naturally looking one. 
Neural networks are also applied for graphics applications. Feed forward networks can learn a mapping between geometric attributes~\cite{Nalbach2017b} or voxels~\cite{phuoc2018rendernet} to shaded outputs, and rerendering networks can correct the artifacts of traditional rendering approaches~\cite{meshry2019neural, Brualla2018}. However, these approaches often produce blurry results and have limited control over the input: appearance changes are mostly restricted to viewpoint or high-level attributes.

In this paper we present a neural network model that learns how to render a scene given voxels as input. The scene can be modified in terms of appearance, location, orientation and lighting, and all changes are faithfully expressed in the rendered output. With only requiring a rough specification of the geometry as a voxel grid, our framework produces an accurate image, with plausible light interactions between the scene elements (\eg casting shadows, reflections, \etc).
Our method includes a rerendering module which enables us to render highly textured objects precisely and in detail.
Moreover, our framework naturally inputs limited appearance information. Instead of manually painting materials on the geometry~\cite{substance} or requiring multiple views of the object~\cite{Seitz2006, unity}, our method can use a single image aligned with a 3D object to capture detailed appearance properties and propagate them to other views.

We demonstrate the ability of our approach to render realistic images in a comprehensive set of experiments. We show geometric and appearance modifications in synthetic datasets with increasing texture complexity and we reproduce the look of objects from real images as well. We illustrate how the network reproduces the interactions between the scene elements, \eg specular reflections, shadows, secondary bounces of light. Finally, we compare our proposed framework with alternative approaches~\cite{phuoc2018rendernet, pix2pix2017, mv3d} and we attain better performance in several metrics.
Our contributions can be summarized as:
\begin{itemize}
    \vspace{-2mm}
    \item A neural rendering framework with controllable object appearance and scene illumination effects.
    \vspace{-2mm}
    \item Capturing texture details with a neural rerendering module.
    \vspace{-2mm}
    \item Learnable interactions between scene entities such as reflections, shadows and secondary bounces of light.
\end{itemize}

%% file: 2_Related.tex
\section{Related Work}
\vspace{-2mm}
\mypar{Geometry-based neural rendering.}
One approach for image synthesis with geometric information is to replace the traditional rendering pipeline with a neural network.
RenderNet~\cite{phuoc2018rendernet} learns how to map a voxel grid to a shaded output, such as Phong shading. The method can also be used for normal estimation, allowing the use of the Phong illumination model~\cite{Phong1975}, and generating textures of faces using PCA coefficients as input. Texture fields~\cite{oechsle2019texturefields} is estimating the appearance of a 3D object using a function that maps a point in space to a color, conditioned on an input image.
In contrast, our approach allows detailed manipulation and rendering of arbitrary textures and provides an adjustable illumination source (area light with soft shadows).

Deferred Neural Rendering~\cite{thies2019neural} learns to synthesize novel views of a scene using neural textures, a learnable element that acts as a $UV$ atlas. Deep Appearance Models~\cite{Lombardi2018} encode the facial geometry and texture of a particular person and can generate novel views during inference. Neural Volumes~\cite{Lombardi2019Volumes} focuses on the 3D reconstruction of an object by taking a set of calibrated views as an input and producing a 3D voxel grid that is rendered with differentiable ray-marching. While these approaches produce high quality results, their models are trained for a particular object/scene, limiting their applicability to general cases, and they assume static light. Also, these methods allow limited edits in the original scene as they focus on view synthesis.

Another direction is neural rerendering. Lookinggood~\cite{Brualla2018} uses a neural network to fix artifacts from a multiview capture and \cite{meshry2019neural} rerenders a point cloud from a 3D reconstruction with modifiable appearance. Similarly,~\cite{pittaluga2019revealing} estimate the RGB image from a structure-from-motion pointcloud. Deep Shading~\cite{Nalbach2017b} converts a set of rendering buffers (position, normals, \etc) to shaded effects. Again, the ability to modify the output is bounded by either what it was presented during training or to holistic appearance transfers.

\mypar{Image-based neural rendering.}
Given a set of images and their corresponding camera matrices, Deep Voxels~\cite{sitzmann2019deepvoxels} encode the view-dependent appearance of an object in a 3D latent voxel grid, which can be later used for rendering novel views. However, the latent voxels incorporate the appearance of a single object that undergoes viewpoint changes and the method requires a large number of calibrated images. 
Another generative approach is to synthesize the images directly given few attributes as an input~\cite{Dosovitskiy15} (\eg transformation parameters, color \etc), but this approach needs access to the whole database of objects and the rendered images are often blurry.

The image-to-image translation~\cite{pix2pix2017} paradigm can also be seen as a form of neural rendering. Methods can synthesize new images of humans based on a 2D pose~\cite{LiuBody, chan2019dance, Liqian17, Balakrishnan2018SynthesizingIO} or convert semantic maps to natural images~\cite{wang2018pix2pixHD, park2019SPADE}. Style transfer~\cite{Gatys2016a} can also alter the appearance of an input image in a realistic way~\cite{Li2018ACS, luan2017deep}. However, these approaches perform specific synthesis and editing tasks, where the control over the final rendering is based on the attributes from the supplied data.

Novel view synthesis is the task of generating a new view of an object given an input image from another view. This can be assisted by a 3D object~\cite{RematasCVPR14}, considering visible and hidden parts~\cite{zhou2016view, tvsn_cvpr2017} or directly from one~\cite{mv3d} or multiple~\cite{olszewski2019tbn, sun2018multiview} views. Again, these methods deal only with rotation modifications and their outputs often lose appearance details. An alternative to the previous explicit geometric transformations are the disentangled representations in feature space. The work of~\cite{Kulkarni15} learns to render different viewpoints and lighting conditions by a careful handling of the training procedure, while~\cite{WorrallInterpretable17} considers the transformations directly on the latent features. Recently,~\cite{NguyenPhuoc2019HoloGANUL} demonstrated the ability to generate realistic images in an unsupervised way using a disentangled latent volumetric representation. However, the modifications are again limited to rotations and simple lighting, while our network handles more complex appearance alterations explicitly on the input voxels.

\mypar{Differentiable rendering.}
To overcome the non-differentiable nature of traditional rendering, previous works introduce a differentiable rasterizer~\cite{DIBR19, liu2019softras, kato2018renderer} or splatting~\cite{YifanDSS2019}, propose a differentiable ray-tracer~\cite{drcTulsiani17, Li2018DMC} or have a differentiable, BRDF-based rendering model~\cite{Deschaintre2018, Li2018MaterialsFM, Liu2017Material, Li2017}.
The focus of these methods is inverse rendering (estimate geometry, materials, \etc), while we are interested in the forward synthesis process.

%% file: 3b_Method.tex
\section{Method}
\label{sec:method}
\vspace{-2mm}
\subsection{Overview}
\label{sec:overview}
\vspace{-2mm}
Our goal is to learn a network that renders a realistic image given a voxelized 3D input.
While there are several types of 3D representations for neural rendering (\eg meshes~\cite{kato2018renderer, liu2019softras}, implicit functions~\cite{oechsle2019texturefields,ParkFSNL19, OccupancyNetworks}, rendering buffers~\cite{Nalbach2017b}), we choose to work with voxels because they are easy to input into a neural network and they provide the flexibility for arranging the scene elements in a natural way (\eg a chair on top of the ground).
The output of our network is a realistic image representing the scene from a particular viewpoint given by the user.

\myfigure{world_voxel_coords}{Scene setup: the object is placed in the world coordinates where it can be rotated and translated.  The light can also be translated and the camera can change elevation. As network inputs, the scene is in camera coordinates and the light position is a 3-dimensional $xyz$ vector.}

\mypar{Scene definition.}
Our scene consists of three elements: the object, the ground and the light. The elements are placed in a bounded 3D world that is observed by a camera at a fixed distance. The object and the ground are represented as voxels and we use an area light as this generates more realistic soft shadows than a point light source.

Our approach supports several editable attributes:
(1) The object can be translated across the $x$ and $z$ axis and rotated around the $y$ axis. Also, we can apply local rigid and non-rigid transformations to the voxels.
(2) The light can be translated in a bounded volume above the ground.
(3) The camera elevation can be adjusted.
(4) The appearance of the object can change. 
For (4), we consider a spectrum of modifications, from painting the object/floor with a single uniform color to applying arbitrary textures to the object.

Apart from the editable parameters, the scene has some fixed attributes. There is ambient light, the color of the light is white, the camera focal length is fixed to 40, the object is diffuse, and the floor is slightly specular. Detailed values of the scene setup can be found in the supplementary material.

The network expects the scene to be in the camera coordinate frame. We first apply the modifications and then we convert the scene from world to camera coordinates.

\subsection{Coloring Voxels}
\label{sec:colorvoxels}
\vspace{-2mm}
\mypar{Manual coloring.}
A straightforward approach  would be to color voxels manually.
\begin{wrapfigure}{r}{0.25\textwidth}
%   \centering
    \includegraphics[width=0.24\textwidth]{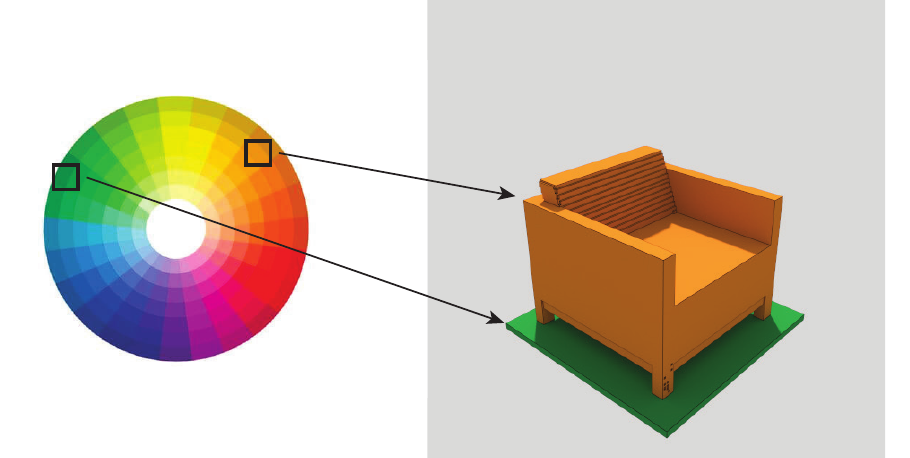}
%   \caption{Birds}
\vspace{-5pt}
\end{wrapfigure}
As a use case we consider the setting where the object and the floor have a single color each.  The user can select the colors from a palette and assign them to the objects in the scene, similar to the bucket fill tool in most image editing software (see inset figure). The network still has to properly shade the scene based on the light position, and generate global illumination effects.

\mypar{Colors from an image.}
Tools for coloring voxels can be found in programs like MagicaVoxel~\cite{magicavoxel}, but detailed voxel painting can be cumbersome. A more practical alternative is to perform image-based coloring. Given the alignment between the 3D object and an image depicting the object (we refer to that image as \textbf{appearance source $\mathcal{A}$}), we can un-project the color of the pixels directly onto the voxels.
\myfigurehere{image_to_voxels}{Coloring voxels from an image: given the 2D-to-3D alignment, the voxels take the color from the corresponding pixels of the appearance source image.}
Alternatively, if the the input 3D object is in the camera space, we can assume orthographic projection. The centers of the voxels are projected to the image using the camera parameters. Finally, the color for every voxel is taken from the pixel it falls into.
While this approach requires aligned images with the 3D objects, recent advantages in 2D to 3D alignment can provide such information automatically, see for example~\cite{RematasPAMI2017, photoshape2018, Aubry14, Wang2016, seeThrough_HuetingReddy_3DV18, Huang2015, Huang18}.

Appearance capture from a single image introduces artifacts typical with projective texturing. The assigned colors come from the input view and they contain view-dependent information such as shading, shadows, \etc. Unlike typical image-based rendering, in our scenario the object will be rendered from a different viewpoint or lighting condition, so the rendered object colors need to change accordingly. We tackle this problem with a carefully prepared training set that includes this type of appearance changes (see~\refSec{experimental-settings}).

Another aspect that requires attention is how to color the voxels that are not visible. We determine voxel visibility automatically using ray-marching~\cite{Levoy1988}. A hidden voxel gets the color of the first visible voxel along its camera ray. We observed that this approach is beneficial in cases with thin structures (\eg chair handles and legs), and the generated artifacts were  well handled by the network.

Additionally, we take advantage of the symmetry in many man-made objects such as chairs and cars: if a voxel is not visible, we copy its color from its symmetric voxel across the $y$ axis (if that is visible).

\subsection{Neural Voxel Renderer (NVR)}
\label{sec:neural-voxel-renderer}
\vspace{-2mm}
\myfigure{network_architecture}{NVR network architecture. Two branches encode the voxel and light position inputs and a decoder combines their output and produces the final rendering.}

\mypar{Model architecture.}
Our Neural Voxel Renderer (NVR) network $\theta$  is illustrated in \refFig{network_architecture} (more details in the supplementary material).
The inputs are
(1) the scene represented as voxels $V \in \R^{128^3 \times 4}$ and
(2) the light position $L \in \R^3$.
The network output the image $I \in \R^{256^2 \times 3}$:
\begin{equation}
    I = \theta (V, L)
\end{equation}
The voxels $V$ contain the RGB colors and visibility, which are automatically estimated based on the 2D-3D alignment (\refSec{colorvoxels}).
We use RenderNet as our backbone~\cite{phuoc2018rendernet}.  The 3D input voxels $V$ are processed by a series of 3D convolutions and a reshaping unit that transforms the 3D features into 2D by reshaping its last two dimensions (\eg $h\times w \times d \times c$ becomes $h\times w \times (d * c)$), followed by $1\times1$ convolutions (\textit{projection unit} in~\cite{phuoc2018rendernet} and \textit{reshape} in \cite{olszewski2019tbn}). The reshaping step can be seen as an orthographic projection in the latent space, with all the depth information being kept. 
The features are then processed by a series of 2D convolutional blocks, leading to the final encoding of the input voxels. 

The light input $L$ is processed by a separate branch with 2 fully connected layers, delivering latent illumination features to the network. These features are then tiled to form an image with the same dimensions as the final features from the voxel encoding branch. In this way, every spatial location in the features has information about the illumination.
After concatenation, the joint features are fed to a decoder that outputs the image in the final resolution.

\mypar{Model training.}
We train the model by minimizing the following loss:
\begin{equation}
    L(I,T) = ||I - T||_1 + \beta \sum_i w_i|| v_i(I) - v_i(T)||_2
    \label{eq:loss}
    \vspace{-2mm}
\end{equation}
where $T$ is the ground-truth target image. The second term is a perceptual loss, with $v_i$ is the response of the $i$ layer of a pretrained VGG~\cite{Simonyan2014VeryDC} network, and $w_i$ its weighting factor. For $i$, we use the \textit{conv1, conv2} layers and set their weights $w_i$ to 1.0 and 0.1 respectively.

The target image $T$ is produced by a traditional, physically-based renderer (Blender Cycles~\cite{blender}) and the object is represented by a 3D mesh. This results in rendering smooth surfaces in image $T$. In this way, the network implicitly learns to map discrete geometric representation such as voxels to a continuous and smooth rendering. 

We train the network using the Adam optimizer~\cite{KingmaB14} with learning rate $10^{-4}$ and a batch size of 10. The 2D convolutional layers are followed by batch normalization and ReLU activations
(more details in supplementary material).

\subsection{Adding a rerendering network (NVR+)}
\vspace{-2mm}
\myfigure{rerender_architecture}{NVR+ network architecture.}

The network in~\refSec{neural-voxel-renderer} is able to render well the overall structure of the scene in terms of colors, reflections, shadows, etc. However, we observe that when the color pattern of the object in the input voxels forms a high frequency and irregular texture, the output is blurry and with artifacts.
For this reason we propose a rerendering network (NVR+) that maintains the high quality textures while producing the correct overall scene appearance (\refFig{rerender_architecture}).

The texture information is already encoded in the voxels' colors and we know the camera parameters of the target rendering (the user sets the scene to be rendered, see~\refSec{overview}). Therefore we can synthesize an image $S$ by splatting~\cite{Zwicker2001SurfaceS} the center of the colored voxels to an empty canvas in the target view. Note that this image will contain the artifacts mentioned in \refSec{colorvoxels} (wrong colors, different shading, \etc) since the target view can be different from the view the appearance was captured from.

The NVR+ consists of three parts: the NVR network described in the previous subsection, the Splatting Processing Network that encodes the synthesized image $S$ into a latent representation, and a Neural Rerendering Network that combines and processes the outputs of the other two networks into the final image. 
The Splatting Processing Network consists of a series of convolutional layers without decreasing the resolution. The output of this network is then added to the features from the NVR and the result is fed to the Neural Rerendering Network. The Neural Rerendering Network processes the combined features with a U-Net~\cite{RonnebergerUnet} architecture and outputs the image in the final resolution. The whole NVR+ network is trained end-to-end, using the same loss as in~\refEq{loss}.

The NVR+ is able to render high-frequency textures accurately and in detail because it combines the best of two modalities. First, the output of the NVR produces a realistic image in terms of reflections, shadows and overall color assignments but it lacks the high-frequency texture details. Second, the output of the Splatting Processing Network contain artifacts from the splatting process, but it also includes features rich in resolution and details. Finally, the Neural Rerendering Network integrates the two network outputs and produces a coherent, detailed and artifact-free final image. The NVR+ renders an image ($256\times 256$) in $\approx 0.1$ sec on a single desktop GPU (Nvidia RTX).

%% file: 4b_Experiments.tex
\section{Experiments}

\subsection{Settings and protocol}
\label{sec:experimental-settings}
\vspace{-2mm}
\mypar{Training.}
For training our models we use 3D shapes from ShapeNet~\cite{shapenet2015} and we render them with Blender Cycles~\cite{blender}, a physically-based path tracer. We focus mainly on the Chairs category, but we also provide qualitative results for the Car category. In all cases the training sets are constructed by sampling randomly 2000 3D objects from the train set as specified in the SHREC’16 challenge~\cite{Savva2016shrec}. Each object is rendered from 20 different viewpoints by uniformly sampling (1) the elevation of the camera, (2) the rotation and translation of the object, and (3) the position of the light.
The camera elevation is between 5 and 50 degrees, the object rotation is uniformly sampled from the 180 degrees hemisphere facing towards the camera, the translation of the object is sampled from an rectangular area around the scene center ($[-0.5, 0.5]$ units) and the light is sampled from a volume above the scene center ($[-1.5, 1.5]$ for the $x$ and $z$ axis and $[2.5, 3]$ for the $y$ axis).
The object is rendered as a mesh for accurate reproduction of its surfaces (giving the target image $T$ in~\refEq{loss}). In contrast, the object is input to the network as a $100^3$ voxel grid, and then it is placed inside the overall $128^3$ voxel scene $V$ to allow rotations and translations (object to world coordinates, see \refFig{world_voxel_coords}).

\mypar{Testing.}
For testing, we want to measure the ability of our model to cope with changes in object rotation and translation, and light translation. We randomly select 40 3D objects from the test set of the SHREC’16 challenge~\cite{Savva2016shrec}. Each object is rendered with the following settings: the camera elevation is randomly sampled between 15 and 45 degrees; a rotation angle range (starting and ending angle) is randomly sampled from the hemisphere facing the camera and the object is rotated with a step of 10 degrees; a start and end location is randomly sampled and the object is being translated between the two end points; similar sampling is applied for the light. This procedure results in 40000 images for training and 1100 images for testing.

\mypar{Appearance settings.}
We generate three settings with varying appearance complexity. \textit{Single color}: where both the object and the floor have a single randomly selected color (RGB values);
\textit{Default}: where the object is rendered with its default ShapeNet textures/materials but the floor color is randomly sampled;
and \textit{Textured}: where the object has a randomly selected texture from the Describable Textures Dataset\cite{cimpoi14describing} (we separate the textures in train and test splits) and the floor has a single randomly selected color.
We train a model for every setting and every category.

\vspace{-2mm}
\subsection{Global illumination effects}
\vspace{-2mm}
Our scenes consists of objects that interact with each other (shadows, reflections, \etc) and our training data was generated with a physically-based renderer. These elements of realism exist in the training dataset, and here we analyze if our network is able to reproduce these effects.

\myfigure{global_illum}{Global illumination effects produced by our framework.}
In \refFig{global_illum} we illustrate how our framework renders global illumination effects.
In the first row we show the reflections on the specular ground for different objects. The overall structure is represented properly and it is following the orientation of the object, even with thin structures.
In the second row we show how our network is rendering shadows; again, thin structures produce thin shadows and concave objects allow the light to pass.
Finally, in the last row we show the effect of multiple light bounces: the color of the object (first two columns) and the ground (last two columns) is affected by the color of the other object. The original color of the object is shown in the small rectangle inset.

\subsection{Neural rendering analysis}
\vspace{-2mm}
In this subsection we analyze the design choices from Section~\ref{sec:method} and evaluate their effects on the final renderings. We additionally compare with alternative techniques for neural rendering and show that our framework performs better both quantitatively and qualitatively. 

\myfigure{single_color}{Results of NVR for the \textit{single color} setting. The voxel colors are properly mapped to pixels in the output images.}

\mypar{Manual coloring.}
In this scenario, we use the \textit{Single color} appearence setting as training data. With this experiment, we want to illustrate the capability of our method (NVR) to map the voxel colors to rendered pixels.~\refFig{single_color} shows that the networks learns to render accurately the object, with correct colors and shading.

\mypar{Colors from image.}
In this scenario, the voxels take their color from an input image (appearance source image $\mathcal{A}$, see~\refSec{colorvoxels}). This is a more challenging setting, as some parts of the object are hidden and the colors contain view-dependent appearance information. Hence, it forms a good test case for evaluating the ability to render more complex appearance (different parts, textures, \etc) while capturing light interactions among scene elements. \refFig{comparison} shows the output of our NVR and NVR+ models in the \textit{default} appearance setting for the Chair category (5th and and 6th column).
The NVR model properly assigns the colors to the individual parts, but fine-grained details on textured areas are washed out.
The NVR+ model renders detailed textures while also accurately producing object shading, ground reflections and shadows.
\refFig{cars} (left) shows the output of NVR+ for the \textit{default} appearance setting of the Car category (using a car-specific model, but with different Car instances for training and testing,~\refSec{experimental-settings}).
In the first row we show the voxel input\footnote{rendered with MagicaVoxel~\cite{magicavoxel} in $4$ seconds each using a ray-tracer.} together with the appearance source image $\mathcal{A}$ from where the colors were taken, in second row our prediction and in third row the ground truth. Again, our method can reproduce accurately the details of the cars; also, by taking advantage of the symmetry, we are able to faithfully reconstruct the hidden side.
Finally, we visualize the output of NVR+ on the \textit{textured} appearance setting (\refFig{cars} right). As the images show, this model can handle high frequency and irregular textures (as well as synthesizing specular reflections and shadows as in \refFig{comparison}).
Moreover, this confirms that our model does not memorize the training data but rather learns to map the input color pattern information into an accurate rendered image. 

\myfigure{comparison}{Visual comparison between our models and different methods (see text for details).}

\myfigure{cars}{Neural rendering of cars and textured objects with NVR+.}

\myfigure{projective_texturing}{Artifacts of projective texturing.\vspace{-5pt}}

\mypar{Comparison to alternative methods.}
We use the Chair category (\textit{default} appearance setting) and for every 3D object, one of the rendered images acts as the appearance source image $\mathcal{A}$. 
We compare against a set of alternatives based on recent works on neural rendering.
Since there is no other method that offers control over geometric, appearance and light edits, we modify them to be comparable.
\\
\textit{Projective texturing}: here we use the alignment between the mesh and the input image to texture the mesh (i.e. estimate its UV map). In \refFig{projective_texturing} we illustrate typical projective texturing artifacts: a) the hidden areas copy the appearance of the visible parts, even when taking symmetry into account (red arrows), and b) light is "baked" into the object appearance, making difficult to render with new lighting.
\\
\textit{Single-to-mutliview}: this method is based on~\cite{mv3d} and consists of a U-net which inputs the appearance source image $\mathcal{A}$, the relative object rotation/translation and light position. This method does not use any 3D information.
\\
\textit{Image encoding}: here we have a setup similar to~\cite{phuoc2018rendernet} but instead of painting the voxels directly, we pass $\mathcal{A}$ through a network to get a latent code that is then appended to the input voxels (similar to how~\cite{phuoc2018rendernet} rendered textured faces). The rest of the network has similar architecture to NVR so the light can be given as an input.
\\
\textit{Image translation}: this method is based on~\cite{pix2pix2017}; the input is the rerendered image $S$ (generated by splatting the voxels onto the image plane) together with the light position as additional channel. The desired output is the target image $T$.
\\
\textit{Deep shading}: here the setup is similar to~\cite{Nalbach2017b}. We use projective texturing to estimate the UV map of the mesh; then we place the mesh in the desired position and we render the diffuse color and depth buffer. We use these buffers together with the light position as inputs to a Pix2Pix~\cite{pix2pix2017} network and we optimize for the target image $T$.

\begin{table}[t]
\small
\centering
\begin{tabular}{ c||c|c|c  }
%  \hline
Method & MSE & DSSIM & Perceptual\\
 \hline
 Single-to-multiview   & 34.8    & 0.040 &   40.17\\
 Image encoding&   24.3  & 0.030   &24.41\\
 Ours (NVR) &22.8 & 0.028&  21.25\\
 \hline
 Image translation    &21.6 & 0.026&  18.12\\
 Deep shading & 21.8& 0.024& 16.54\\
 Ours (NVR+)&   \textbf{14.3}  & \textbf{0.012}&\textbf{8.21}\\
%  \hline
\end{tabular}
\vspace{2pt}
\caption{Evaluation of different methods for neural rendering.\vspace{-5pt}}
\label{tab:eval}
\end{table}
Table~\ref{tab:eval} compares the performance of the different methods on mean squared error (MSE), structural dissimilarity (DSSIM) and the perceptual loss in \refEq{loss} (using the same layers $i$ and weights $w_i$).
NVR+ performs better than the alternatives in all metrics by a large margin. This illustrates the ability of the rerendering module to reproduce fine details, especially for textured 3D objects.
\refFig{comparison} provides a qualitative comparison of the different methods. As can be seen, \textit{Single-to-mutliview} results are blurry, while \textit{Image encoding} captures the overall color but fail to assign it correctly to the individual parts. The \textit{Image translation} model produces typical GAN artifacts and cannot estimate the shadows properly. The \textit{Deep shading} model also faces similar limitations, despite using a mesh representation instead of voxels.
Our NVR model captures the color and structure of the scene, but smooths out the fine texture details.
Finally, our NVR+ model accurately renders both the geometry and the texture of the object, while also realistically synthesizing shadows, specular reflections and highlights in the output image.

\mypar{Voxel resolution.} The object has an initial voxel resolution of $100^3$ and is then placed in a scene $V\in 128^{3\times4}$. Here, we investigate the rendering quality when the initial voxel resolution varies. In Table~\ref{tab:resolution} and~\refFig{resolution} we show the performance of smaller resolutions for the NVR+ model (which are then rescaled with nearest neighbor interpolation). The performance decreases gracefully and even at a very low resolution ($25^3$) our method produces plausible outputs.
\begin{table}[t]
\small
\centering
\begin{tabular}{c|cccc}
% \hline
& $100^3$& $75^3$ & $50^3$ & $25^3$\\
\hline
MSE & 14.3& 14.4& 15.3& 19.3\\
DSSIM & 0.012& 0.014& 0.016& 0.024\\
Perceptual & 8.21& 9.47& 10.9& 18.68\\
% \hline
\end{tabular}
\vspace{2pt}
\caption{Effect of voxel resolution on performance.\vspace{-25pt}}
\label{tab:resolution}

\end{table}

\vspace{-4pt}
\subsection{Editing analysis}
\vspace{-2mm}
\mypar{Illumination edits.}
When the light source changes position, the scene appearance should change accordingly. In \refFig{shading}, we illustrate this effect: as the light sources moves, the brightness of different parts of the object changes and shadows/light reflections in the ground move accordingly.

\myfigure{resolution}{Rendering an object with different voxel resolutions.}
\myfigure{shading}{Illumination effects by changing the light position. Our framework changes properly the overall shading (e.g. the back of the chair is brighter in position 2) and the shadows.}
\myfigure{Appearance_edits}{Applying geometric (left) and appearance (right) modifications.}
\myfigure{realistic}{Neural rendering with natural illumination as an input.\vspace{-4pt}}
\mycfigure{pix3d}{Rendering real objects. In the first row there are the appearance source images and in second and third row renderings of the objects using the NVR+ network.\vspace{-5pt}}
\myfigurehere{other_categories}{Rendering different categories. The categories of these real objects were not part of the training dataset (trained only on the chair category).}

\mypar{Object geometry edits.}
Apart from global object rotations and translations, we can also deform the object. In \refFig{Appearance_edits} left we visualize the effect of scaling across an axis, resulting in elongated or squeezed versions of the object.

\mypar{Object appearance edits.}
Detailed modifications on the appearance source image can propagate through our neural renderer. In this example, we manually paint patterns and letters on the appearance source image, so during the coloring step (\refSec{colorvoxels}) the edits pass on to the voxels.
\refFig{Appearance_edits} right shows how our method can synthesizes images with the object in new viewpoints and the light in new positions while preserving these fine-grained edits that were made to the appearance source image.

\mypar{Increasing realism.} In this experiment, we investigate the use of more realistic ways to illuminate the scene. We modify the NVR+ network so instead of the $xyz$ light coordinates, it takes as an input an $32\times32$ environment map. The environment map is processed by a series of convolutional layers to extract a latent code, which is then supplied to the NVR+ network.
We additionally consider a textured circular ground and add specularity to the \textit{default} dataset objects.
We use 80 environment maps for training and 20 for testing, taken from \cite{hdrhaven}. Results are shown in \refFig{realistic}, with the appearance source image shown in an inset.

\mypar{Appearance capture from real images.}
So far we have experimented with synthetic images with varying appearance complexity. However, our framework can capture object appearance from any input image. In this experiment, we use the Pix3D dataset~\cite{pix3d} which contains aligned pairs of images with 3D objects. We use the real image as the appearance source and we map it to to the voxels of the provided 3D object as before. Note that unlike the previous experiments with ShapeNet objects, Pix3D also includes scanned objects with imperfect geometry.
In \refFig{pix3d} we present our renderings when the voxels and the appearance come from real images. Our framework is able to faithfully render these objects despite not being trained on real objects and despite significantly different geometric, illumination and material conditions than the training set.

\mypar{Testing on other categories.}
Our framework extends to other categories that were not included in the training dataset. In this experiment we take the NVR+ network trained on the default Chair category and we apply it to real images from other categories.
In \refFig{other_categories} we illustrate the rendering for sofas, tables and other miscellaneous objects from Ikea~\cite{lpt2013ikea} and Pix3D~\cite{pix3d} datasets.

%% file: 5_Conclusion.tex
\section{Conclusion}
\vspace{-6pt}

We presented Neural Voxel Renderer, a framework that synthesizes realistic images given object voxels as input, and provides editing functionalities to the output. Our framework can reproduce the detailed appearance of the input due to a rerendering module that handles high-frequency and complex textures. 
We show a wide range of rendering scenarios, where we modify the input scene with respect to illumination, object geometry and appearance. Moreover, we demonstrate the appearance capture and rendering of real objects from several categories. We hence believe that our neural renderer is a useful tool that advances the state-of-the-art and can spawn further research.